\documentclass{article}
\usepackage{spconf}
\usepackage{amsmath}
\usepackage{graphicx}
\usepackage{cite}
\usepackage{comment}
\usepackage{soul}
\usepackage{xcolor} 
\usepackage{hyperref}
\usepackage{amssymb}

\newcommand\blfootnote[1]{
    \begingroup
    \renewcommand\thefootnote{}\footnote{#1}
    \addtocounter{footnote}{-1}
    \endgroup
}

\definecolor{highlightyellow}{HTML}{FEFF66}
\definecolor{textgreen}{HTML}{688F50}
\definecolor{textorange}{HTML}{CB703A}

\DeclareRobustCommand{\colorgreen}[1]{\textcolor{textgreen}{#1}}
\DeclareRobustCommand{\colororange}[1]{\textcolor{textorange}{#1}}
\DeclareRobustCommand{\hlyellow}[1]{{\sethlcolor{highlightyellow}\hl{#1}}}

\title{Reinforcing Pre-trained Models Using Counterfactual Images}
\name{Xiang Li$^\dag$, Ren Togo$^{\dag\dag}$, Keisuke Maeda$^{\dag\dag\dag}$, Takahiro Ogawa$^{\dag\dag}$, Miki Haseyama$^{\dag\dag}$}
\address{$^\dag$Graduate School of Information Science and Technology, Hokkaido University, Japan\\
    $^{\dag\dag}$Faculty of Information Science and Technology, Hokkaido University, Japan\\
    $^{\dag\dag\dag}$Data-Driven Interdisciplinary Research Emergence Department, Hokkaido University, Japan\\
    E-mail:\{xiang\_li, togo, maeda,  ogawa, mhaseyama\}@lmd.ist.hokudai.ac.jp}
\begin{document}
\ninept
\maketitle
%
\begin{abstract}
This paper proposes a novel framework to reinforce classification models using language-guided generated counterfactual images. Deep learning classification models are often trained using datasets that mirror real-world scenarios. In this training process, because learning is based solely on correlations with labels, there is a risk that models may learn spurious relationships, such as an overreliance on features not central to the subject, like background elements in images. However, due to the black-box nature of the decision-making process in deep learning models, identifying and addressing these vulnerabilities has been particularly challenging. We introduce a novel framework for reinforcing the classification models, which consists of a two-stage process. First, we identify model weaknesses by testing the model using the counterfactual image dataset, which is generated by perturbed image captions. Subsequently, we employ the counterfactual images as an augmented dataset to fine-tune and reinforce the classification model. Through extensive experiments on several classification models across various datasets, we revealed that fine-tuning with a small set of counterfactual images effectively strengthens the model. \blfootnote{This work was partially supported by the JSPS KAKENHI Grant Numbers JP21H03456, JP23K11141 and JP23K11211.}

\end{abstract}
\section{Introduction}
\label{sec:intro}
Deep Neural Networks (DNNs)~\cite{schmidhuber2015deep} have become powerful tools for pattern recognition, excelling at extracting complex patterns from visual data. However, the ``black box" nature of these models undermines their performance, casting doubt on their reliability~\cite{castelvecchi2016can}. The intricate, multi-layered, and non-linear structure of DNNs makes their decision-making processes opaque, rendering the models less trustworthy and verifiable in real-world applications. This lack of transparency poses a significant obstacle to the widespread acceptance and adoption of these models by users.

The ``black box" issue of DNN's decision-making mechanisms presents various challenges. Especially when the consequences of misclassification are significant, understanding the rationale behind a model's output is crucial. For instance, in the creative industries involved in image generation and style transformation~\cite{xu2018attngan}, the unpredictability of generative models often leads to outcomes that deviate from user expectations. Therefore, to enable the generation of intended images, it is essential to understand and further improve the model's performance. Efforts in this direction are known as Explainable AI (XAI)~\cite{gunning2019xai}, which is an area of active research.


XAI aims to provide explanations for models' decision-making processes~\cite{goyal2019counterfactual,dhurandhar2018explanations,vandenhende2022making}. A common approach within XAI involves visualizing areas of focus within the model by displaying intermediate layer outputs as heatmaps. This allows for a visual understanding of the regions the model is concentrating on, thereby helping to comprehend the model's characteristics. However, even with these approaches, certain aspects remain opaque, such as why the model focuses on specific areas and the impact on other classes. A recent technique being explored is the generation of visual counterfactual explanations~\cite{vandenhende2022making}. This approach seeks to present counterfactual explanations that visually depict scenarios that could not actually occur, gaining particular attention in fields like image classification and image generation~\cite{augustin2022diffusion}.


However, the use of counterfactual images in classification tasks is primarily aimed at providing explanations, rather than actionable solutions for addressing identified weaknesses. Although these methods effectively highlight dependencies and biases within models, they often lack a definitive strategy for enhancing model robustness. For example, attention heatmaps and feature visualizations can clarify certain aspects of model decisions, but they infrequently translate into direct improvements in model performance. Moreover, in efforts to enhance model performance, existing approaches frequently do not meet expectations. Similarly, basic data augmentation strategies might fail to accurately simulate the complex scenarios encountered in real-world applications, resulting in generic enhancements that do not address specific issues.


In this paper, we propose a novel model reinforcing method based on the Language-guided Counterfactual Images (LANCE)~\cite{prabhu2023lance} framework to address the aforementioned issues. Our method aims to overcome the limitations of existing counterfactual explanation techniques and enhance the robustness of classification models. By generating counterfactual images based on natural language descriptions, our approach not only identifies the model's weaknesses but also utilizes explanations obtained from stress tests to expose vulnerabilities. Moreover, by integrating these customized counterfactual images into the training set, the proposed method enables the model to learn from past misclassifications and fortifies its weaknesses. This approach not only clarifies the causes of the model's weaknesses but also systematically addresses them, leading to a more robust and reliable classification system. Extensive experiments using the HardImageNet dataset have validated the effectiveness of our framework. Our method significantly improves both the interpretability of models and their classification robustness, thereby enhancing the flexibility of approaches to the diversity and unpredictability of real-world data. It can contribute to the societal application of deep learning as a novel method for model reinforcing.

The contributions of this paper are summarized as follows.
\begin{enumerate}
\item \textbf{Development of the Language-guided Counterfactual Image (LANCE) Framework:}
We propose a novel model reinforcing method based on language-guided counterfactual images to overcome the limitations of existing visual counterfactual explanation methods. Our approach enables the identification and improvement of robustness in classification models using counterfactual images generated based on language.

\item \textbf{Utilization of Counterfactual Images for Model Reinforcement:} Our method not only identifies weaknesses in the model but also utilizes insights from counterfactual images to fortify it. By integrating customized counterfactual images into the training set, the model learns from its past misclassifications, thereby becoming more robust.

\item \textbf{Improved Adaptability to Real-world Data:} Through rigorous experiments on the HardImageNet dataset, we validated the effectiveness of our framework, which uses language-guided counterfactual images. These experiments demonstrated significant improvements in both the classification accuracy and interpretability of the models.
\end{enumerate}

\vspace{-5pt}
\section{Related work}
\label{sec:related}
In this section, we focus on visual counterfactual explanations in computer vision, the generation of counterfactual images, and the robustness of deep learning models.

\vspace{-5pt}
\subsection{Evolution of Visual Counterfactual Explanation}
The concept of visual counterfactual explanations has experienced a remarkable transformation, starting with the research of adversarial examples that were well-designed to test the robustness of classification models. Previous studies have introduced subtle yet strategically designed substitutions, revealing the weakness that seemingly tiny modifications can mislead predictions of models~\cite{szegedy2013intriguing, goodfellow2014explaining}. The explanations gained from these initial tries have encouraged further research, prompting the development of more sophisticated substitutions. Unlike prior methods, the substitutions are not only aimed at triggering misclassification. They are carefully edited to maintain the semantic consistency between counterfactual images and the original, for example, they only substitute the semantically same parts of images. As a result, these counterfactual images are contextually relevant and can provide more practical explanations~\cite{goyal2019counterfactual, vandenhende2022making}.

As the field has been developed, the emphasis has shifted toward generating counterfactual images that preserve the underlying context and content of the image, while still leading to a misclassification from the model~\cite{stepin2021survey}. This meticulous approach to counterfactual generation helps to better understand the hidden layers in neural networks, offering a more transparent view into their intricate decision-making processes~\cite{dhurandhar2018explanations}. By maintaining most parts of the original image's caption and just perturbing one word or phrase, these advanced counterfactual models serve as a more effective tool for testing the classification model's logical ability. It enables a deeper understanding of the key features and patterns that influence its judgments. This evolution indicates a move towards counterfactual images that challenge the models' robustness. It also enriches the comprehension of the model's interpretative framework, marking a significant step forward in the quest for XAI~\cite{alicioglu2022survey}.

\subsection{Exploring Decision Networks with Generated Counterfactual Images}
The exploration of generative models as tools for indicating the inner workings of visual models has become an increasingly important field of research. A widely adopted strategy in this field involves the generation of minimal image perturbations that are capable of changing the model’s predictions. Before generating, this strategy makes subtle changes such as altering the curvature of a subject's lips to influence the output of a ``smiling" classifier~\cite{goyal2019counterfactual}. Our research follows this idea to provide perturbations to image regions that are outlying to the ground truth of the classification task. For instance, we study whether variations in features like hair color can swing a classifier's judgment of a swimming cap. Thus, we can reveal the model's implicit biases and see if the unexpected weight is placed on irrelevant features.

Luo \textit{et al}.~\cite{luo2023zero} is similar to our approach, wherein they operate a weighted combination of edit vectors within the latent space of StyleGAN~\cite{karras2019style} to alter model predictions while preserving the global structure and semantics of the image. Despite the novelty of this method, it is limited by the need to pre-define the attributes to be edited. This may reduce its applicability to more diverse and complex datasets. Our methodology, when following the intent of Luo \textit{et al}.~\cite{luo2023zero}, seeks to overcome these limitations by adopting a more flexible approach without having a detailed understanding of the dataset to counterfactual image generation.

Our work is also in line with the efforts of Li \textit{et al}.~\cite{li2023imagenet}, who use diffusion models to construct robustness benchmarks. These are characterized by a variation of background and object attributes. This approach to robustness benchmarks is invaluable in assessing a model's resilience to a controlled set of attribute alterations. Furthermore, Wiles \textit{et al}.~\cite{wiles2022discovering} generate the root cause of misclassifications through cluster analysis, providing subtle conditions under which models are easy to make mistakes. All of these studies contribute to a more accurate understanding of model behaviors. This opens the way for the development of more sophisticated XAI systems.

\begin{figure*}[t!]
    \centering
    \includegraphics[keepaspectratio, width=16cm]{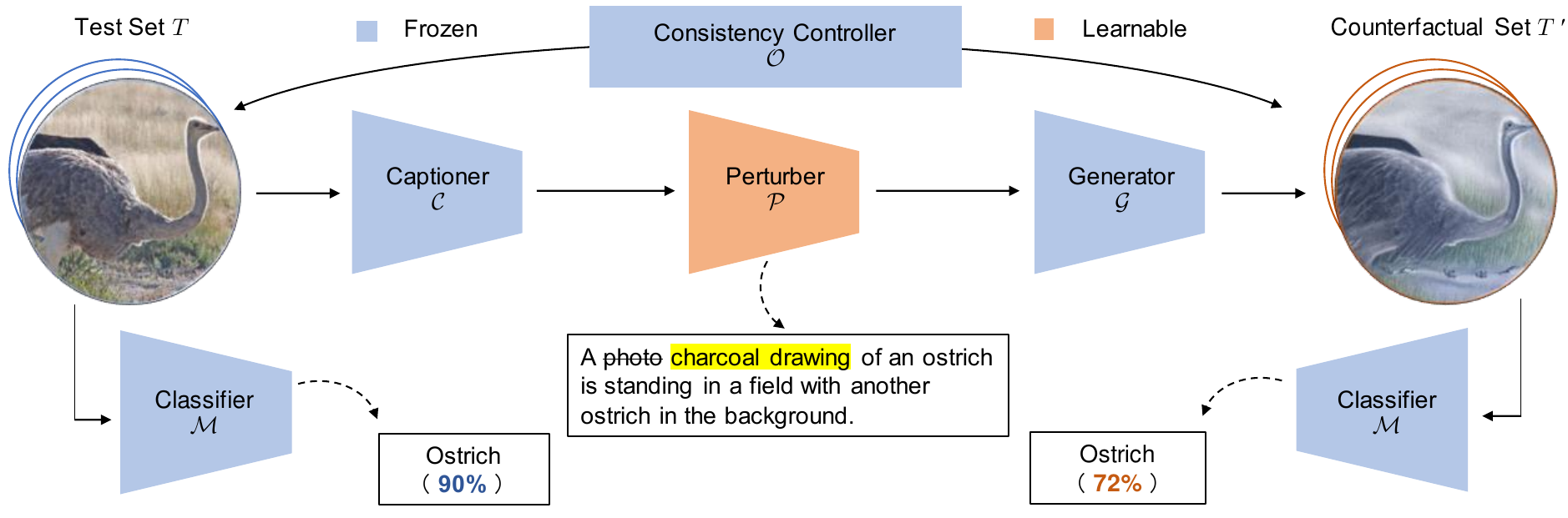}
    \caption{\textbf{Overview of the weakness identification process.} 
    We train a model for generating counterfactual examples to assess the robustness of classification models. This approach uses a pre-trained model and a test dataset to produce image descriptions with a captioning model, which are then modified by a Large Language Model (LLM) to change their original meaning. The modified captions and images are used for counterfactual image editing through a text-to-image generator, applying various perturbations to create a test set that extensively evaluates the model's robustness. Finally, the model's sensitivity and accuracy changes are thoroughly assessed using this test set.
    }
    \label{fig:wfo}
\end{figure*}

\subsection{Language-guided Counterfactual Image Generation}
The introduction of natural language processing in image generation has made for the emergence of novel methodologies for counterfactual image generation. The Language-guided Counterfactual Image Generation (LANCE) framework~\cite{prabhu2023lance} stands at the forefront of this intersection, using the descriptive power of language prompts to guide the generation of counterfactual images. This breakthrough approach has enabled the generation of counterfactual images to differ from the original images in a meaningful way. Additionally, it also maintains a level of visual and semantic coherence that resonates with human cognition. The counterfactual images produced by LANCE are not subjective but filled with clearness, reflecting the subtlety provided by the guiding text descriptions.

Although the LANCE framework represents a significant advance in generating counterfactual images, our research aims to extend these boundaries by exploring previously unaddressed challenges. Our goal is twofold: first, to reveal and emphasize model weakness by the generated counterfactual images, and second, to use these images to guide the model training process to directly address the weakness and reduce misclassification. We achieve this by deeply integrating insights from counterfactual explanations into the model's learning framework. Specifically, we aim to strengthen the model's ability to identify and prioritize significant features and distinguish them from spurious correlations. For instance, a model might incorrectly learn to associate the presence of snow in an image with the subject being a polar bear, while in reality, the snow is just a common background element for various subjects in the dataset. To counteract this, we expose the model to a diverse set of images where polar bears are depicted in non-typical settings, such as grasslands or even deserts. This approach encourages the model to focus on the defining characteristics of polar bears rather than the coincidental presence of snow, leading to a more accurate and robust understanding. By integrating this process, we can test the present comprehension of the model and further direct its development towards a more reliable XAI system.

\begin{figure*}[t!]
    \centering
    \includegraphics[keepaspectratio, width=16cm]{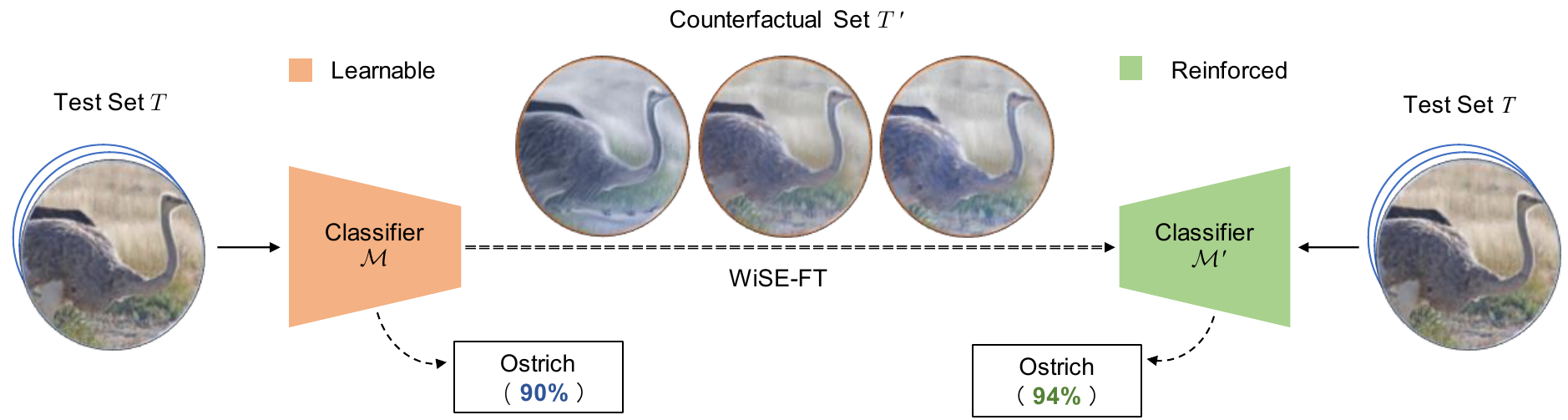}
    \caption{\textbf{Overview of model reinforcement.} Following the identification of a specific weakness in the classification model, we start a process using a counterfactual image dataset targeting this weakness. The classifier after being fine-tuned with this targeted dataset enhances its flexibility. The impact of this reinforcement is then verified against a test set, confirming the model's improved robustness and accuracy.}
    \label{fig:wro}
\end{figure*}

\section{Proposed Reinforcing Method}
In this study, we introduce a novel framework that strategically applies language-guided counterfactual images to reinforce and refine classification models. A fundamental idea of our methodology is that counterfactuals can be used not only to evaluate the model but also to improve it. This process begins with the identification of model weaknesses through a series of stress tests using counterfactual images (see Fig.~\ref{fig:wfo}). Once these weaknesses are pinpointed, we enter the crucial phase of model reinforcement, wherein the model is fine-tuned with a targeted set of counterfactuals based on the identified weaknesses (see Fig.~\ref{fig:wro}). This targeted reinforcement approach ensures that the model not only gains a deeper understanding of complex visual patterns but also becomes stronger at maintaining robust performance in the face of various situations in the real world.

\subsection{Perturber Model Fine-tuning}
\label{pmft}
The start of our method lies in the generation of counterfactual images that are closely linked to textual captions. As shown in Fig.~\ref{fig:wfo}, we first introduce a pre-trained BLIP-2~\cite{li2023blip} as the captioner $\mathcal{C}$. To encourage the generation of descriptive captions with minimal redundancy, we configure the captioning process with a minimum limit of 20 words and enforce a significant repetition penalty. This configuration is designed to promote the production of detailed and varied textual captions that effectively guide counterfactual image generation. Subsequently, we employ a Large Language Model (LLM) as the perturber $\mathcal{P}$ to edit the generated captions. To train a perturber part that is capable of generating caption alterations,  we decide on five different visual variation factors for the stress test.

\begin{enumerate}
\item \textbf{Subject:} Alterations to the image caption's subject (e.g., from ``man" to ``woman" or from ``football player" to ``basketball player") test the model's recognition accuracy under the presence of diverse or atypical subjects, some of which may be absent from the training data.
\item \textbf{Object:} Changes to the object in the caption (e.g., from ``apple" to ``orange" or ``pear") challenge the model's robustness by introducing new associations between concepts.
\item \textbf{Background:} Adjustments to the background within a caption (e.g., from ``mountain" to ``beach" or ``ice sheet") test the model's ability to generalize scenes across a range of environments and weather conditions. 
\item \textbf{Adjective:} Variations in descriptive terms (e.g., from ``old" to ``new" or ``young") examine the model's adaptability to the full range of attributes that define an object's appearance.
\item \textbf{Data domain:} Shifts in the domain of content displayed (e.g., from ``photo" to ``sketch" or ``paint") assess the model's flexibility across different artistic representations and data distributions.
\end{enumerate}
While this selection is not exhaustive, it covers a broad range of visual variability. Building on this foundation, we implement Low-Rank Adaptation (LoRA) for fine-tuning a LLAMA-7B model~\cite{touvron2023llama}, serving as the perturber $\mathcal{P}$. Throughout the fine-tuning process, we maintain the original LLAMA-7B model in a frozen state to ensure that its foundational structure remains unaltered. The adaptation is confined to the weight matrices $ W_\mathcal{P} \in \mathbb{R}^{d \times k}$ within the self-attention modules, where post-adaptation changes are constrained to a low-rank structure. Specifically, the fine-tuned weight matrix $W_\mathcal{P}^{ft}$ is expressed as
\begin{equation}
\label{eq:pft}
W_\mathcal{P}^{ft} =  W_\mathcal{P}^{pt} + \Delta W_\mathcal{P} = W_\mathcal{P}^{pt} + \textbf{AB},
\end{equation}
with $ \textbf{A} \in \mathbb{R}^{d \times r}$, $ \textbf{B} \in \mathbb{R}^{r \times k}$, and $W_\mathcal{P}^{pt}$ means the pre-trained weight. By maintaining a low rank $r$, we ensure that the number of trainable parameters in matrices $\textbf{A}$ and $\textbf{B}$ remains minimal, allowing for efficient adjustments to the model's behavior. 

Furthermore, to preserve the significance of counterfactual images, we avoid edits that alter the ground truth class in captions. For example, if ``carrot" is the correct label, edits changing ``carrot" straightly to ``turnip" are filtered out. Employing sentence BERT~\cite{devlin2018bert}, we evaluate the semantic similarity between the original and edited captions, filtering out those edits that are too closely aligned with the ground truth caption. This ensures that the applied perturbations remain relevant without changing the original intent.

\subsection{Classification Weakness Identification}
\label{cwi}
When obtaining the fine-tuned perturber model, we utilize a text-to-image latent diffusion model, specifically Stable Diffusion~\cite{rombach2022high}, as the generator $\mathcal{G}$ to generate counterfactual images that are conditioned on the original image. To achieve directed image editing, we introduce the prompt-to-prompt technique, which strategically modifies the cross-attention maps in accordance with the caption edits during denoising diffusion steps. 

Moreover, we incorporate the null-text inversion technique from Mokady \textit{et al}.~\cite{mokady2023null} to refine our image generation process. For an original input image $x$ with its caption $c$, we start with its encoded latent vector $z_0$ and reverse the diffusion for $K$ steps ($z_0 \rightarrow z_K$) using the Denoising
Diffusion Implicit Model (DDIM)~\cite{song2020denoising} approach. Our model employs classifier-free guidance, running the diffusion with and without text prompts. To accurately reconstruct the original image, we adjust $\emptyset_k$, the null-text embedding at each timestep $k$, to minimize the mean square error between the predicted latent code $\hat{z}_k$ and the initial diffusion state $z_k$, ensuring the reverse diffusion path remains close to the original encoding $z_0$. Let $S_{k-1}(\hat{z}_k, \emptyset_k, c)$ denotes one single step of deterministic DDIM sampling. The optimization process is defined as follows:
\begin{equation}
\label{eq:nti}
\min_{\emptyset_k}{\left \| z_{k-1} - S_{k-1} (\hat{z}_k,\emptyset_k,c)  \right \|^2_2  }. 
\end{equation}
However, the efficiency of image editing via prompt-to-prompt with null-text inversion hinges on a critical hyperparameter, denoted as $\tau $, which decides the diffusion steps during which the self-attention maps from the original image are utilized. This hyperparameter $\tau $ is fine-tuned to match the degree of the desired edit; for instance, more weighty alterations, such as those to the background or weather conditions, typically require a smaller $\tau $ value. We adopt an automated component, as the consistency controller $\mathcal{O}$, to standardize this hyperparameter. This controller $\mathcal{O}$ involves iterating over a predefined range of $\tau$ values and selecting the optimal one based on the CLIP~\cite{radford2021learning} directional similarity metric~\cite{gal2022stylegan}. The metric evaluates the alignment of changes in the original image $x$ and corresponding caption $c$ within the embedding space. The image and text encoders of CLIP are represented by $\mathcal{O}_I$ and $\mathcal{O}_T$, respectively. The directional similarity criterion of CLIP, denoted as $\mathcal{O}$, is defined as follows:
\begin{equation}
\label{eq:ec}
\mathcal{O}= 1 - \frac{(\mathcal{O}_{I}(x)-\mathcal{O}_{I}(x')) \cdot (\mathcal{O}_{T}(c)-\mathcal{O}_{T}(c'))}{ \left |\mathcal{O}_{I}(x)-\mathcal{O}_{I}(x')\right |  \cdot \left | \mathcal{O}_{T}(c)-\mathcal{O}_{T}(c')\right | },
\end{equation}
where $x'$ means the counterfactual image and $c'$ presents the perturbed captions. Utilizing this metric, we achieve a balance, ensuring that the generated counterfactual image is meaningfully aligned with the edited caption while remaining within a reasonable divergence from the original image in the embedding space of CLIP.

Finally, we assess the performance of classifier model $\mathcal{M}$ by monitoring the classification accuracy across both the original test set $T$ and the generated counterfactual test set $T'$. Our focus is on quantifying the reduction in top-5 accuracy (Acc@5) on $T'$, relative to $T$, as a measure of the model's robustness to the introduced perturbations. Denote the ground truth class as $y$, the metric for this comparison is defined as follows:

\vspace{-5pt}
\begin{equation}
\label{eq:deltaacc}
\begin{split}
\Delta \mathrm{Acc@5} = &\left [  \frac{1}{\left | T' \right | }\sum_{(x',y)\in T'} \mathrm{Acc@5}(\mathcal{M} (x'),y)  \right ]\\
                    &- \left [  \frac{1}{\left | T \right | }\sum_{(x,y)\in T} \mathrm{Acc@5}(\mathcal{M} (x),y)  \right ]. \\
\end{split}
\end{equation}

\subsection{Process of Model Reinforcement}
\label{pwtr}

Based on the foundation laid by the fine-tuned perturber model and the subsequent identification of classification weaknesses, we advance to the critical phase of targeted reinforcement. We suggest a strategy to turn the identified weakness into practical insights that can directly influence the model's training solutions. As shown in Fig.~\ref{fig:wro}, the goal is to systematically reinforce the classifier model $\mathcal{M}$, enhancing its resilience to the specific weaknesses exposed through the counterfactual test set $T'$. During this procedure, there exists a substantial problem of causing catastrophic forgetting which is a phenomenon where a model loses its learned ability such as to accurately classify other classes. To ease this risk, we are inspired by the method of Weight-Space Ensembles for Fine-Tuning (WiSE-FT)~\cite{wortsman2022robust}.

In practice, we use a selective fine-tuning approach where parameters of most parts of the model, layers other than the classification head, are kept frozen to preserve the pre-existing knowledge base. We then fine-tune the classification head with a blend of newly learned and reserved parameters, adjusting only a fraction, denoted by the hyperparameter $\alpha$, while the remaining $1 - \alpha$ portion of the parameters are conserved from the original model. We formalize this update process of parameter $U_\mathcal{M'}$ as $U_\mathcal{M'}((\mathcal{M'}(x),y), (1 - \alpha) \cdot \theta_0 + \alpha \cdot \theta_1)$, where $\theta_0$ represents the original model parameters, $\theta_1$ is the fine-tuned parameters and $\mathcal{M'}$ means the reinforced classification model. The hyperparameter $\alpha$ is adjusted based on classification accuracy $(\mathcal{M'}(x),y)$. This strategy enables us to integrate the new counterfactual data without overriding the model's established competencies. 

\begin{figure}[t!]
    \centering
    \includegraphics[keepaspectratio, width=8cm]{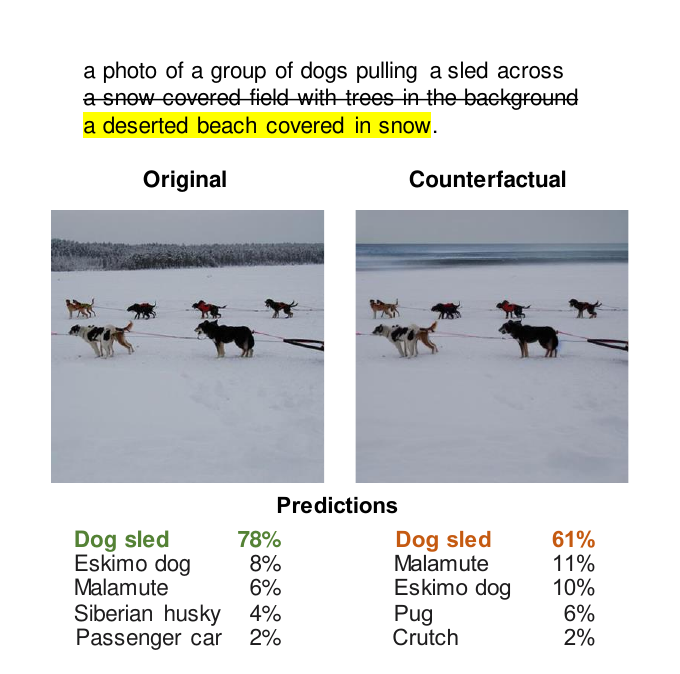}
    \vspace{-10pt}
    \caption{\textbf{Visualization examples of counterfactual images.} We present the array of original and generated counterfactual images. For each image pair, the top-5 predicted classes determined by a trained VGG-16 are shown, along with the corresponding predicted probability. The \st{strikethrough} is original words while the \hlyellow{highlighting} is perturbed words. The correct predictions are denoted in \textbf{\colorgreen{green}} for original images and \textbf{\colororange{orange}} for counterfactual images.
}
    \label{fig:cie}
    \vspace{-10pt}
\end{figure}

By leveraging the counterfactual image dataset, which specifically targets the model's weakness, we conduct a focused recalibration of the classification head. This targeted adjustment is carefully calibrated to ensure that the model not only overcomes its previous limitations but also emerges with an enhanced ability to handle the diverse challenges presented by real-world data.

\section{Experiments}
\label{sec:exp}
To validate the effectiveness of our proposed framework, we conducted several experiments. We detail the dataset, model selection, evaluation metrics, and the comprehensive analysis in the following subsections.

\vspace{-5pt}
\subsection{Experimental Setup}
To train our perturber model, we employed GPT-3.5 turbo~\cite{brown2020language} and programmatically handle a compact dataset of approximately 600 caption perturbations, including the visual variation factors mentioned in Sec.~\ref{pmft}. For example, to achieve the ``adjective" modification, we used the following prompt to generate counterfactual edit: \textit{``Generate all possible variations of the provided sentence by only adding or altering a single adjective or attribute."} This automated process was applied to a random selection of captions from the MSCOCO dataset~\cite{lin2014microsoft}.

\begin{table}[t!]
 \caption{\textbf{Result of weakness identification.} We identified various models which are pre-trained on ImageNet-1K. The counterfactual test sets increase the difficulty of classification as demonstrated by the decline in the performance of models.}
 \label{table:cfi}
 \centering
 \resizebox{\columnwidth}{!}{
  \begin{tabular}{ccccc}
   \hline
    \textbf{Test Set}  &  \textbf{Class} &  & \textbf{Acc@5}($\uparrow$) \\
    \cline{3-5}
    \rule{0pt}{2.5ex}
    & &  ResNet50~\cite{he2016deep} & DenseNet121~\cite{huang2017densely} & VGG16~\cite{simonyan2014very}  \\
    \hline \hline
    \rule{0pt}{2.5ex}
     & Dog sled & 95.43 & 96.55 & 82.76\\
   HardImageNet & Howler monkey & 82.67 & 79.80 & 83.74\\
   Test Set $T$& Seat belt & 86.29 & 85.71 & 81.28 \\
     & Ski & 78.22 & 71.43 & 69.46 \\
    \hline 
    \rule{0pt}{2.5ex}
     & Dog sled & 80.77 & 82.27 & 61.54\\
   Counterfactual & Howler monkey & 40.39 & 44.33 & 48.28\\
   Test Set $T'$& Seat belt & 69.14 & 72.66 & 69.78 \\
     & Ski & 68.72 & 58.59 & 44.33 \\
   \hline\hline
    \rule{0pt}{2.5ex}
     & Dog sled & -14.66 & -14.28 & -21.22\\
    Reduction in& Howler monkey & -42.28 & -35.47 & -35.46\\
    Accuracy($\mathrm{\Delta Acc@5}$)& Seat belt & -17.15 & -13.05 & -11.5 \\
     & Ski & -9.5 & -12.84 & -25.13 \\
    \hline
  \end{tabular}}
\end{table}

\begin{figure*}[t!]
    \centering
    \includegraphics[width=0.8\linewidth]{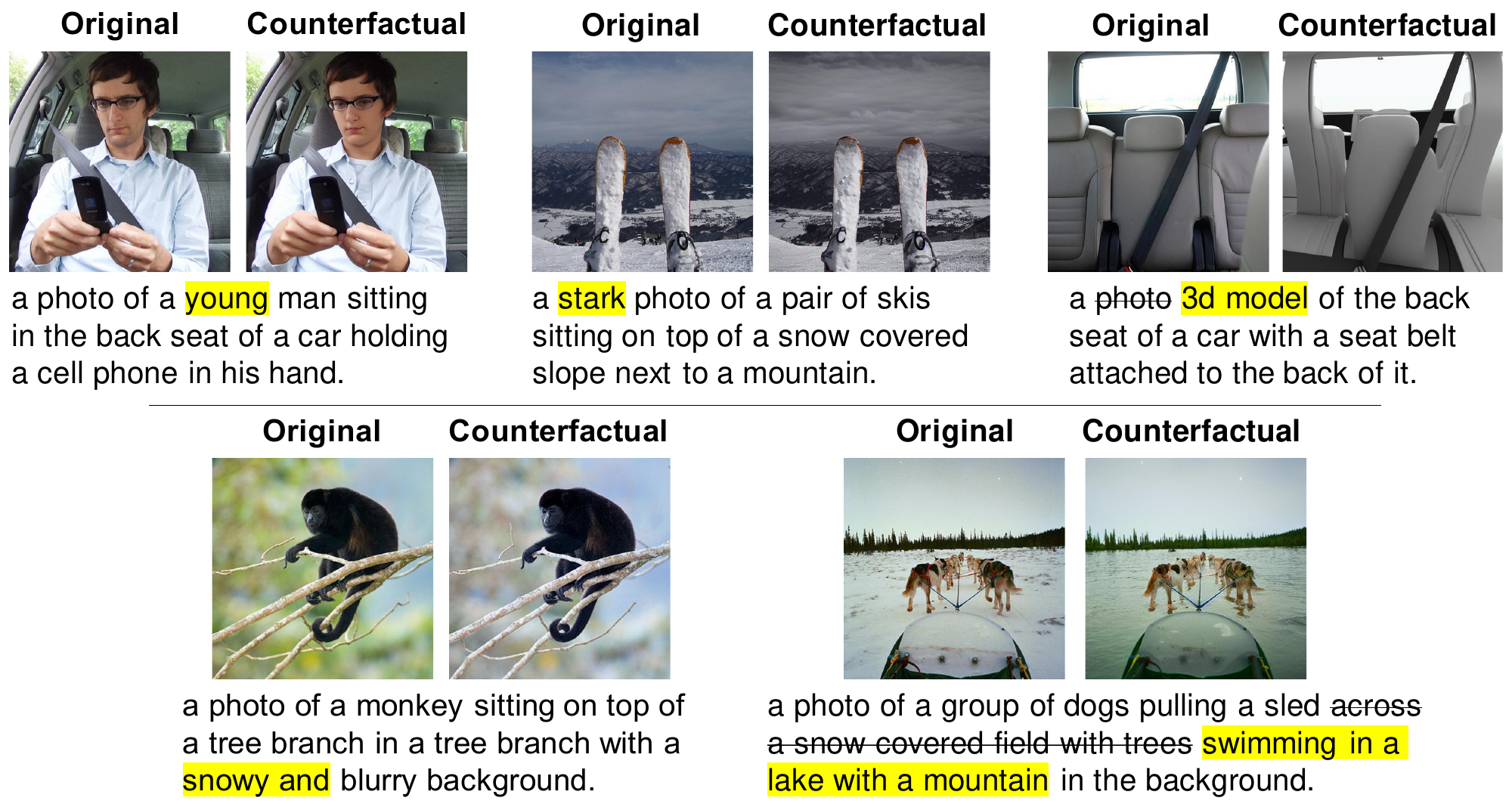}
    \vspace{-5pt}
    \caption{\textbf{Result example of generated counterfactual images.} The \st{strikethrough} is original words while the \hlyellow{highlighting} is perturbed words.}
    \label{fig:wie2}
    \vspace{-5pt}
\end{figure*}

\begin{table*}[t!]
 \caption{\textbf{Result of model reinforcement.}  We conduct the comparative analysis of model performance on the HardImageNet~\cite{moayeri2022hard}, OOD, and hybrid test sets. ``Baseline" means original models trained by ImageNet-1K, ``Standard" shows models fine-tuned by standard data augmentation dataset, and ``Counterfactual" presents models targeted reinforced by our method. We use \textbf{bold} to indicate the highest accuracy while \underline{underline} represents the second-highest.}
 \label{table:cfr}
 \centering
 \resizebox{\textwidth}{!}{
  \begin{tabular}{cc|ccc|ccc|ccc}
   \hline
    &  & \multicolumn{9}{c}{\textbf{Acc@5}($\uparrow$)} \\
    \cline{3-11}
    \rule{0pt}{2.5ex}
    \textbf{Test Set} & \textbf{Class} & & ResNet50~\cite{he2016deep} &  & & DenseNet121~\cite{huang2017densely} &  & & VGG16~\cite{simonyan2014very}  \\
    \cline{3-11}
    \rule{0pt}{2.5ex}
    & &  Baseline & Standard & Counterfactul &  Baseline & Standard & Counterfactul &  Baseline & Standard & Counterfactul \\
    \hline \hline
    \rule{0pt}{2.5ex}
     & Dog sled & \underline{95.43} & \textbf{96.64} & 96.20 & \textbf{96.55} & 95.57 & \underline{96.06} & 82.76 & \underline{84.73} & \textbf{87.19}\\
   HardImageNet & Howler monkey & 82.67 & \underline{84.37} & \textbf{85.33} & 79.80 & \underline{82.27} & \textbf{84.24} & 83.74 & \textbf{86.21} & \underline{84.73}\\
   Test Set & Seat belt & \underline{86.29} & 85.22 & \textbf{87.19} & 85.71 & \underline{87.19} & \textbf{87.68} &81.28 & \underline{84.24} & \textbf{86.70}\\
     & Ski & 78.22 & \underline{78.82} & \textbf{83.25} & 71.43 & \textbf{72.94} & \underline{71.91} & \underline{69.46} & \textbf{71.28} & 68.10\\
    \hline 
    \rule{0pt}{2.5ex}
     & Dog sled & \textbf{84.29} & 78.57 & \textbf{84.29} & 85.14 & \underline{86.20} & \textbf{89.71} & 74.29 & \textbf{78.57} & \underline{75.71}\\
   OOD & Howler monkey & 78.25 & \underline{80.00} & \textbf{88.75} & 84.73 & \underline{85.21} & \textbf{88.25} & \underline{60.00} & 56.25 & \textbf{63.75}\\
   Test Set & Seat belt & 69.95 & \underline{70.52} & \textbf{72.20} & 71.92 & \textbf{75.71} & \underline{72.48} & \underline{48.57} & 47.14 & \textbf{52.86}\\
     & Ski & 81.54 & \textbf{83.08} & \textbf{83.08} & 75.08 & \underline{75.38} & \textbf{76.92} & 63.08 & 63.08 & \textbf{64.62}\\
   \hline 
   \rule{0pt}{2.5ex}
     & Dog sled & \underline{88.33} & 85.83 & \textbf{89.17} & 88.33 & 88.33 & \textbf{91.67} & 78.83 & \textbf{82.17} & \underline{80.50}\\
     Hybrid& Howler monkey & 82.13 & \underline{83.08} & \textbf{87.69} & 80.04 & \underline{83.58} & \textbf{86.15} & \underline{67.54} & 66.69 & \textbf{69.85}\\
   Test Set & Seat belt & 75.33 & \underline{75.67} & \textbf{77.01} & 75.12 & \textbf{80.23} & \underline{77.50} & 66.50 & \underline{67.67} & \textbf{69.83}\\
     & Ski & 80.48 & \underline{81.35} & \textbf{83.12} & 73.43 & \underline{73.98} & \textbf{74.10} & 66.09 & \textbf{67.02} & \underline{66.96}\\
   \hline
   \end{tabular}}\textbf{}
   \vspace{-5pt}
\end{table*}

Next, we selected the HardImageNet~\cite{moayeri2022hard} validation set, which comprises 15 ImageNet classes known for relying on false features, as the original dataset to generate counterfactual images. This dataset is particularly suited for assessing our counterfactual reinforcement effectivenes. However, since the counterfactual images are generated based on HardImageNet, to ensure a comprehensive and objective assessment, we prepare a new Out of Distribution (OOD) test set for the evaluation. 
The OOD dataset consists of real-world images that are more difficult to classify. It probably provides more challenging images than HardImageNet, which better evaluates algorithmic performance in realistic settings. Furthermore, we synthesized a hybrid test set, random incorporation of images from both HardImageNet and the OOD datasets, to provide a reliable assessment of our model's reinforcement performance.

As classifier models, we selected ResNet50~\cite{he2016deep}, DenseNet121
~\cite{huang2017densely}, and VGG16~\cite{simonyan2014very} due to their diverse architectural paradigms and widespread adoption in the field. We conducted comparative experiments using models pre-trained on ImageNet-1k to determine a baseline. We also used models that were fine-tuned on a dataset enhanced with standard data augmentation techniques. These techniques included random horizontal and vertical flipping, random masking, random rotation, random cropping, and random adjustments to brightness, contrast, and saturation.

For our evaluation metric, we employed Acc@5 to capture a wider range of top predictions. During the reinforcement experiments, we fine-tuned the models with a batch size of 8 and a learning rate of 0.0001, applying early stopping with a maximum of 50 epochs and a min$\Delta$ of 0.005 to prevent overfitting. Additionally,  we set the hyperparameter $\alpha$ to 0.3, balancing the weights to optimize performance without reducing robustness.

\subsection{Results and Discussion}
As shown in Fig.~\ref{fig:cie} and Table~\ref{table:cfi}, we observed a consistent decrease in performance across various ImageNet-1K pre-trained models when evaluated against the counterfactual test sets. This performance degradation demonstrates the greater classification challenges created by counterfactual images, validating the role of counterfactual images in revealing the model weakness. Moreover, Fig.~\ref{fig:wie2} illustrates examples of generated counterfactual images. It is evident that language-guided counterfactual image generation can maintain the main semantic information of images unchanged while editing other parts of images. Consequently, they extremely produce challenging images with captions for classification tasks. This not only aids in enhancing the interpretability of the model but also helps visually comprehend the weaknesses of the model.
Next, according to Table~\ref{table:cfr}, our method demonstrated a clear advantage over the baseline and standard fine-tuned models. This trend was evident across the HardImageNet, OOD, and hybrid test sets, indicating a robust reinforcement of the models' capabilities. Therefore, we verify the effectiveness of the proposed framework.

In the current phase, our study has several limitations. While this paper proposes a novel approach, the comparison is still limited to standard data augmentation techniques, and future research should explore a broader range of data augmentation and fine-tuning methods. Including other generative counterfactual image methods for a more comprehensive analysis is also significant. Moreover, incorporating a wider set of metrics could offer a more holistic evaluation of the reinforcement effectiveness. Additionally, the realism of counterfactual image generation requires refinement, as some counterfactual edits may not accurately reflect real-world scenarios, and implementing more subtle or abstract perturbations poses challenges. Addressing these areas will improve the effectiveness and applicability of our approach.

\section{Conclusion}
\label{sec:conc}
We have introduced a novel counterfactual reinforcement framework designed to overcome weaknesses in classification models. Through a meticulous process of generating language-guided counterfactual images, identifying classification weaknesses, and addressing these weaknesses, we demonstrated that our approach could significantly enhance the robustness of models. The experimental results confirmed the superiority of our method, which consistently outperformed both baseline models and standard data augmentation techniques in accuracy across various challenging datasets. The successful application of our framework to widely recognized classification models indicates that our method is broadly applicable and can serve as a foundation for future research aimed at developing reliable XAI.


\vfill\pagebreak

\bibliographystyle{IEEEbib}
\bibliography{ICIP2024}
\end{document}